\documentclass{article}
\usepackage{ijcai20}

\usepackage{times}

\usepackage{soul}
\usepackage{url}
\usepackage[hidelinks]{hyperref}
\usepackage[utf8]{inputenc}
\usepackage[small]{caption}
\usepackage{graphicx}
\usepackage{amsmath}
\usepackage{booktabs}
\usepackage{color}
\usepackage{multirow}
\usepackage{lscape}
\usepackage{graphicx}
\title{Towards information-rich, logical text generation with knowledge-enhanced neural models}

\author{
Hao Wang$^1$\and
Bin Guo$^1$\footnote{Contact Author}\and
Wei Wu$^{2}$\And
Zhiwen Yu$^1$\\
\affiliations
$^1$Northwestern Polytechnical University, Xi’an, China\\
$^2$Microsoft Research, Beijing, China\\
\emails
wanghao456@mail.nwpu.edu.cn,
\{guob, zhiwenyu\}@nwpu.edu.cn,
wuwei@microsoft.com
}

\begin{document}

\maketitle

\begin{abstract}
Text generation system has made massive promising progress contributed by deep learning techniques and has been widely applied in our life. However, existing end-to-end neural models suffer from the problem of tending to generate uninformative and generic text because they cannot ground input context with background knowledge. In order to solve this problem, many researchers begin to consider combining external knowledge in text generation systems, namely knowledge-enhanced text generation. The challenges of knowledge-enhanced text generation including how to select the appropriate knowledge from large-scale knowledge bases, how to read and understand extracted knowledge, and how to integrate knowledge into generation process. This survey gives a comprehensive review of knowledge-enhanced text generation systems, summarizes research progress to solving these challenges and proposes some open issues and research directions.
\end{abstract}

\section{Introduction}
Text generation, also known as natural language generation (NLG), aims to make machines express like humans, which has the capability to produce smooth, meaningful and informative textual contents. From the original template-based and statistical methods to the deep learning-based methods, text generation has attracted the attention of massive researchers and made many remarkable advances. Depending on the data sources, text generation can be divided into \textit{text-to-text}, \textit{data-to-text}, and \textit{image-to-text} generation. We focus on the text-to-text generation in this survey, because it still has massive research challenges and a wider range of applications. Text-to-text generation takes natural language text as input, understands the input text to obtain semantic representations, and generates corresponding output text according to task requirements.

\begin{figure*}
	\centering
	\includegraphics[scale=0.5]{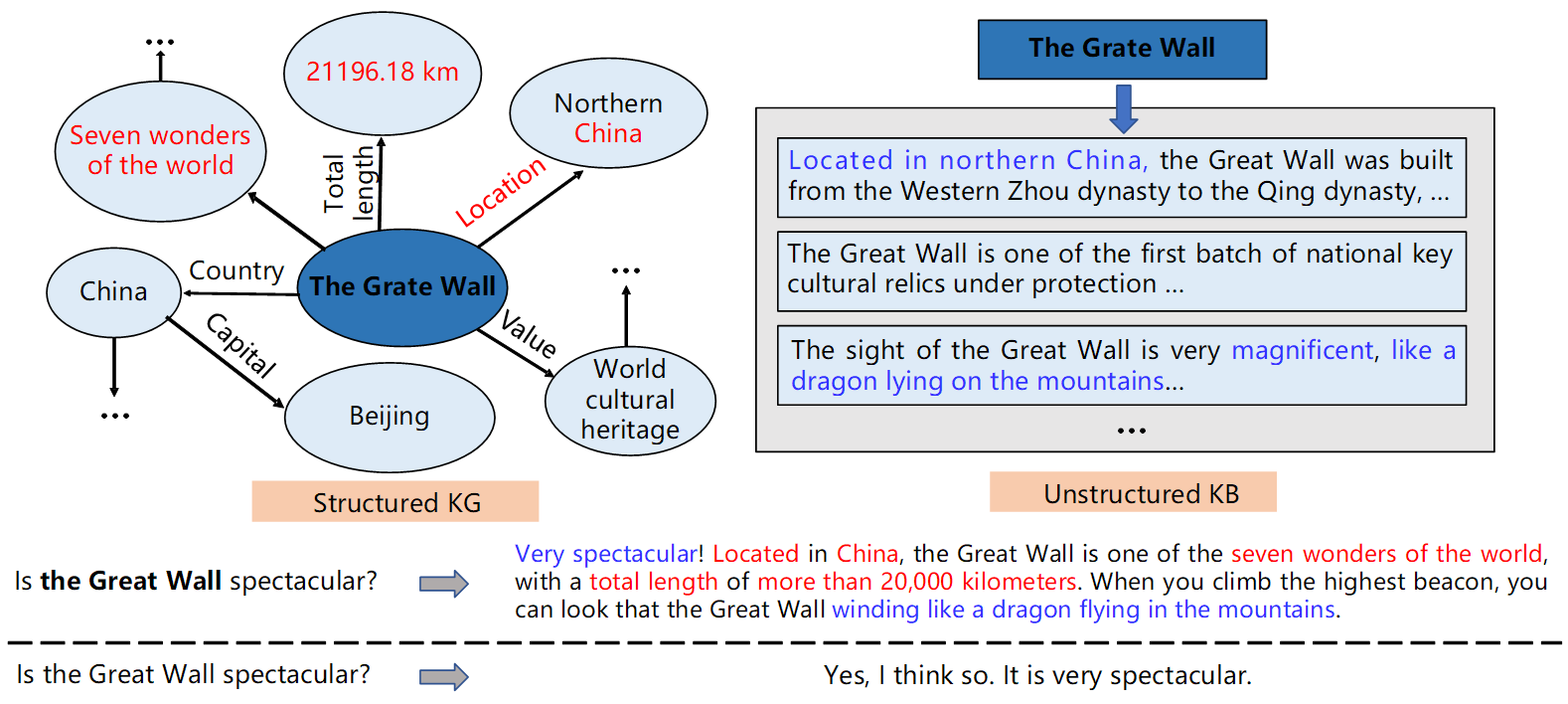}
	\caption{Two dialogue examples with (the first line) or without(second) combining external knowledge}
	\label{figure1}
\end{figure*}

Most advances in text generation in recent years are benefited from deep neural networks, such as recurrent neural network (RNN) \cite{elman1990finding} and Transformer \cite{vaswani2017attention}. With the help of these technologies, text generation systems have been able to generate smooth, topic-consistent and even personalized text. However, existing text generation systems lack interactions with the real world, and have little access to the external knowledge, making them easy to generate the short and meaningless text. We humans are constantly acquiring, understanding and storing knowledge and will automatically combine our knowledge to understand the current situation in communicating, writing or reading, which is a huge challenge faced by text generation systems. To generate more informative,diverse and logical text, text generation systems must have the ability to combine external knowledge, which is a promising research direction. Fig.~\ref{figure1} is an example of a dialogue system with or without commonsense knowledge, where we can see that combining with commonsense knowledge, including structured knowledge graph (KG) composed of triples and unstructured knowledge base (KB) composed of natural language text, the dialogue agent can generate more informative, diverse and logical response.

There are many researchers from both academia and industry have begun to explore knowledge-enhanced text generation system by incorporating different types of knowledge. Due to the complexity of the real world, the scale of knowledge base is usually extremely large. How to extract the most relevant knowledge from massive knowledge facts based on the simple text input is a huge challenge because of the diversity of natural languages. Meanwhile, how to effectively understand the extracted knowledge and integrate it into neural network models to facilitate text generation is also a difficult problem. To our best knowledge, this is the first survey to summarize knowledge-enhanced text generation systems in detail. To sum up, we summarize contributions of our work as follows.

\begin{itemize}
\item We briefly introduce the development process of text generation, formalize the definition of knowledge-enhanced text generation, and analyze a number of key challenges in this research area.
\item We summarize the current research progress in knowledge-enhanced text generation systems by systematically categorizing the state-of-the-art works according to research challenges.
\item We propose some open issues and research directions for the reference of the community.
\end{itemize}

\section{Background and formalized definition}
In this section, we briefly introduce the development of text generation, formalize the definition of general and knowledge-enhanced text generation, and summarize some research challenges.

\begin{table*}[]
\resizebox{\textwidth}{!}{%
\begin{tabular}{|c|l|l|}
\hline
\multicolumn{1}{|l|}{\textbf{Knowledge types}} &
  \textbf{Challenges} &
  \textbf{Existing solutions} \\ \hline
\multirow{2}{*}{Structured KG} &
  Representing knowledge graphs with vectors &
  \begin{tabular}[c]{@{}l@{}}Word embedding-based knowledge representation \cite{wang2019knowledge}\\ Distance-based knowledge representation \cite{gunel2019mind}\\ Graph attention-based knowledge representation \cite{guan2019story}\\ \end{tabular} \\ \cline{2-3}
 &
  Incorporating knowledge vectors into models &
  \begin{tabular}[c]{@{}l@{}}Concatenated with input vector \cite{liu2018knowledge}\\ Attention-based knowledge graph decoder \cite{qiu2019machine}\\ GCN-based knowledge incorporation \cite{de2019question}\end{tabular} \\ \hline
 \multirow{2}{*}{Unstructured KB} &
  Extracting knowledge &
  \begin{tabular}[c]{@{}l@{}}Keyword matching \cite{ghazvininejad2018knowledge}\\ Semantical level knowledge extraction \cite{lian2019learning}\end{tabular} \\ \cline{2-3}
 &
  Reading and understanding knowledge &
  \begin{tabular}[c]{@{}l@{}}Memory network-based knowledge understanding \cite{madotto2018mem2seq}\\ Transformer-based knowledge understanding \cite{Kim2020Sequential}\\ RL-based knowledge understanding \cite{xu2019end} \end{tabular} \\ \hline
\end{tabular}%
}
\caption{A summary of challenges in knowledge-enhanced text generation system}\label{tab:1}
\end{table*}

\subsection{NNLM and RNNLM}
The neural network language model (NNLM) \cite{bengio2003neural} is firstly proposed for text generation tasks, which leverages neural networks to model language representation. NNLM maps the input into a low-dimensional space, thereby reducing the parameters of the model. Given the text sequence $s_{n} = [\omega _{1},\omega _{2},...,\omega _{n}]$ and a $\theta$-parametrized language model $L_{\theta }(\omega |context) = \hat{P}(\omega|context)$, NNLM can be approximated as Eq.~\ref{NNLM}, where $\omega_{t}$ represents the t-th word in $s_{n}$.
\begin{equation}
\hat{P}(\omega_{t}|context) \approx  P(w_{t}|\omega_{t-n+1},\omega_{t-n+2},...,\omega_{t-1})
\label{NNLM}
\end{equation}

NNLM has achieved promising results in experiments, but as a typical feedforward neural network, the length of the context it receives must be fixed in advance, so it cannot capture context information of variable length. To solve this problem, the RNN language model (RNNLM) \cite{mikolov2010recurrent} is proposed, which is an auto-regressive language model that utilizes RNN to encode variable-length inputs into vector representations. This process can be formulated as Eq.~\ref{RNNLM}.
\begin{equation}
\hat{P}(\omega_{t}|context) \approx  P(w_{t}|RNN(\omega_{1},\omega_{2},...,\omega_{t-1}))
\label{RNNLM}
\end{equation}

Theoretically, RNNLM can model text sequences of any length. However, due to the fixed dimension hidden vector, the information in excessively long context may not be stored efficiently. Therefore, variants of RNN, including long short-term memory (LSTM) and gated recurrent unit (GRU), have been utilized to improve the performance of RNNLM, which perform well in capturing long-term dependencies.

\subsection{Encoder-decoder framework}
The length of input and output text of above language models are equal but there are many cases where the length of them are different, e.g. the question and answer in the QA system. To deal with this problem, the Encoder-decoder framework \cite{sutskever2014sequence} is proposed, composed of an encoder and a decoder. The encoder uses RNN to encode the input sequence $X=(x_{1},x_{2},...,x_{m})$ into the intermediate semantics representation $c$, where $m$ is the length of input sequence. The decoder utilizes another RNN to generate the t-th output word $y_{t}$ according to $c$ and $y_{1},y_{2},...y_{t-1}$. This process can be defined as Eq.~\ref{encdec}, where $Y=(y_{1},y_{2},...,x_{n})$ and $n$ is the length of the output text sequence.
\begin{equation}
p(Y|X) = \prod_{t=1}^{T}p(y_{t}|X,y_{<t})
\label{encdec}
\end{equation}

Encoder-decoder is a very general computing framework, whose specific model in the encoder and decoder can be adjusted based on tasks. It has been widely used in various text generation tasks since proposed, such as machine translation, text summarization and dialogue system.

\subsection{Knowledge-enhanced text generation}
In addition to the Encoder-decoder framework, various neural network models and algorithms have been applied to text generation tasks to improve the quality of generated text. The development of text generation systems requires the generated text to be more informative, diverse and logical rather than simple smooth or surface correct. Combined with external knowledge, text generation systems can deeply understand the input, and generate more informative, more consistent with the logic of human expression, and with less common sense mistakes. Given a set of knowledge facts $F = \{f_{i}\}_{i=1,2,...,k}$, where each $f_{i}$ may be a text sequence or a knowledge triple, and $k$ is the number of facts, the knowledge-enhanced text generation model can be formulated as Eq.~\ref{knowledge}.
\begin{equation}
p(Y|X,F) = \prod_{t=1}^{T}p(y_{t}|X,F,y_{<t}).
\label{knowledge}
\end{equation}

There are two forms of external knowledge, that is structured KG and unstructured KB. The KG is essentially a semantic network containing multiple types of entities and relations with the form of $<head,relation,tile>$, where $head$ and $tile$ are different entities. Entities refer to things in the real world and relations express connections between entities. The KB has no fixed form and usually store knowledge related to specific concepts in textual sequence form. There are many challenges in combining external knowledge into text generation systems, as shown in Table~\ref{tab:1}. When combining structured knowledge, the first challenge is how to obtain vector representations of knowledge triples as acceptable inputs to neural network models. Neural network models need input data with vector form, while the information stored in structured KB is symbolized. It is a difficult problem to map these symbols into low-dimensional dense vector spaces. And how to incorporate knowledge vectors as additional input into neural network models to guide the generation process is also a challenge. Because knowledge facts in the unstructured KB are stored with the form of natural language text, mapping knowledge facts to vector representations does not pose a research challenge. The first challenge in combining unstructured knowledge is how to extract the most appropriate knowledge from massive knowledge facts due to the possible semantic duplication of different knowledge. The understanding of sentence-level natural language text is a long-term research challenge in NLP, so how to efficiently read and understand textual knowledge facts to integrate them into generation systems is another challenge in combining unstructured knowledge. Researchers have made great efforts to address these challenges, which will be detailed summarized in following sections.

\section{Text generation with structured KG}
Structured KGs can store a wider range of knowledge types but less information due to its simple representation of triples. However, its unique symbolic storage form is quite different from vectors required by neural network models. Therefore, how to map knowledge triples into low-dimensional vector representations and efficiently incorporate knowledge vectors into neural network models are key directions of the research, which will be summarized in this section.

\subsection{Word embedding-based knowledge representation}
The simplest way to obtain vector representations of structured KBs is to directly treat entities and relations in knowledge triples as common words, and then use word embedding methods to obtain vector representations, which has been widely used at the initial research stage.

In order to comprehensive understand the content of document in reading comprehension tasks, Mihaylov \textit{et al.} \cite{mihaylov2018knowledgeable} utilize a BiGRU to encode knowledge triples as text sequences to get the key-value memory. And then the Key-Value retrieval algorithm selects a single sum of weighted fact representations for each token to enhance the understanding of the document context. Wang \textit{et al.} \cite{wang2019knowledge} propose a KB-based single-relation QA system. The entity linking module determines the optimal subject in the question to select knowledge facts, which will be encoded into vectors using a BiLSTM. The relation detection module calculates similarity scores of each question and its relation candidates to select the triple with highest score to answer the question.

\subsection{Distance-based knowledge representation}
There is a gap between KG of symbolic form and vector representation, so directly encoding entities as common words may lead to certain information loss. The concept of knowledge representation learning is proposed to represente entities and relations in low dimensional dense vector spaces for calculation and reasoning. Bordes \textit{et al.} \cite{bordes2013translating} propose the TransE algorithm, which uses the translation invariant phenomenon of word vector and distance-based scoring function to obtain vector representations of entities and relations which can be better integrated with text generation system to provide more powerful knowledge support.

Moussallem \textit{et al.} \cite{moussallem2019augmenting} incorporate external knowledge into machine translation system to improve the quality of results. Knowledge facts are linked based on the translated document and encoded by the modified TranE, and then concatenated vectors into the internal vectors of NMT embeddings as the input of the decoder. Gune \textit{et al.} \cite{gunel2019mind} incorporate entity-level knowledge from knowledge graph into Transformer encoder-decoder architecture to produce coherent summaries. Extracted entities are initialized with the pretrained TransE algorithm to get the vector representations and then are fed into separate multi-head attention channel for generating summaries.

\subsection{Graph attention-based knowledge representation}
To obtain more accurate vector representations, the graph attention algorithm \cite{zhou2018commonsense} is proposed, which uses relation information to aggregate entities to generate new entity representations. The attention mechanism makes better use of the interconnections between graph entities and distinguishes the hierarchy of connections, which can enhance the effective information needed in text generation tasks.

To generate more informative responses, Zhou \textit{et al.} \cite{zhou2018commonsense} propose the static graph attention mechanism to generate a static representation for a retrieved graph to augment the semantics of input words. The dynamic graph attention mechanism is designed for attentively reading all knowledge triples for text generation. Guan \textit{et al.} \cite{guan2019story} propose an incremental encoding scheme for story ending generation to mine context clues hidden in the story context and adopt graph attention and contextual attention respectively to obtain the graph vectors. The multi-source attention mechanism combines commonsense knowledge for facilitating story comprehension to generate coherent and reasonable story endings.

For comprehensive understanding paragraph-level document, Qiu \textit{et al.} \cite{qiu2019machine} construct sub-graphs for entities to capture the structural information in the KB. Representations of nodes in a sub-graph are updated using the graph attention network with the document context, which are combined with document representations to generate final answer.

\subsection{Concatenating knowledge with input vector}
After vector representations of knowledge triples are obtained, the next challenge facing by knowledge-enhanced text generation systems is how to integrate knowledge vectors into neural network models. The simplest method is to directly concatenate knowledge vectors with input vectors to enhance vector representations of the input, and then send them into the decoding stage for text generation.

For instance, Young \textit{et al.} \cite{young2018augmenting} extract triples according to entity keys in the input to form a text sequence, which is encoded by LSTM to obtain vector representations. Then knowledge vectors are added with the input vector to calculate the degree of correlation with the alternative responses. Liu \textit{et al.} \cite{liu2018knowledge} propose the idea of entity diffusion which means that conversation usually drifts from one entity to another. The similarity between extracted entities and other entities is calculated to retrieve relevant entities. Word vectors of entities and relations are averaged to obtain vector representations of each triple, which is concatenated with the input vector to guide the response generation.

\subsection{Attention-based knowledge graph decoder}
The attention mechanism can focus on the important contents among the numerous input information and select the key information while ignoring other unimportant. Through the attention mechanism, knowledge-enhanced text generation systems can focus on most critical parts of knowledge, instead of feeding all the selected knowledge directly into neural networks, to produce more informative text.

For example, Moon \textit{et al.} \cite{moon2019opendialkg} construct KG embeddings to represent entities with TransE algorithm and aggregate input contexts with relevant entities. To generate candidate KG entities efficiently, an attention-based graph decoder is proposed to walk an optimal path in a large KG to select candidate entities.

For paragraph-level essay generation, Yang \textit{et al.} \cite{yang2019enhancing} present a memory-augmented neural model to combine commonsense knowledge. Knowledge concepts are extracted using input topics as query and stored into a memory matrix. The model will attend on the memory and dynamically update it to incorporate information of the generated text for diverse and topic-consistent essay generation. Koncel \textit{et al.} \cite{koncel2019text} introduce a graph transforming encoder to leverage the relational structure of knowledge graphs to encode knowledge graphs into vectors and then the decoder will attend on the input title and knowledge graphs to generate informative and topic-coherent text.

\subsection{GCN-based knowledge incorporating}
GCN \cite{kipf2016semi} is a natural extension of CNN in the graph domain which learns node feature and structure information in the end-to-end manner. It is a very powerful neural network framework on graphs so it has begun to attract researchers' attention in text generation systems combining structured knowledge graphs.

De \textit{et al.} \cite{de2019question} consider question answering as an inference problem on a graph of the document collection. Nodes in the graph are entities appeared in the document and edges in the graph represent the relations between entities. The GCN is used to capture reasoning chains by propagating local contextual information along edges to perform multi-step reasoning for generating answers. Lv \textit{et al.} \cite{lv2019graph} extract evidence from knowledge graph and make predictions based on the evidence. The graph-based contextual word representation learning module is used to re-define the distance between words for learning better contextual word representations using graph structural information. The graph-based inference module is applied to encode neighbor information into the representations of nodes using GCN and aggregate evidence to generate answers.

\section{Text generation with unstructured KB}
Unstructured KBs are composed of natural language text related to concepts, which express rich semantic information. Because of its textual form, the unstructured KB can be easily combined with text generation systems whose input is text sequences. However, the scale of knowledge base is usually extremely huge, which contains too redundant information. Therefore, how to extract the knowledge required by text generation systems and efficiently understand the knowledge to integrate it into the generation process are main research challenges. There have been many researches of knowledge-grounded text generation with unstructured KB, which will be discussed in detail in this section.

\subsection{Key word matching-based knowledge extraction}
The simplest way to extract knowledge from unstructured KB is the key matching method using words in the input as keywords. This method is simple and direct, but can only extract knowledge according to the surface information of words, and cannot combine deeper semantic information into the knowledge extraction.

For instance, Ghazvininejad \textit{et al.} \cite{ghazvininejad2018knowledge} firstly introduce external knowledge into the fully data-driven neural conversation model. Given the dialogue history, relevant knowledge facts are identified by keyword matching method using entities in the context as keys. Then retrieved know facts are fed into the memory network to retrieve and weight facts based on the input and dialogue context to enhance the semantic representation of the input.

\subsection{Semantical level knowledge extraction}
The simple key word matching method may make it hard to accurately select the required knowledge due to the less information contained in single word. Therefore, many researchers focus on the knowledge selection in the semantic level and put forward many novel ideas.

The same query in human conversation may be related to different responses, so different knowledge may be utilized. To solve this problem, Lian \textit{et al.} \cite{lian2019learning} propose the idea of the posterior distribution over knowledge, which is calculated from both the input query and response to provide more accurate guidance on knowledge selection. By minimizing the distance between the prior and the posterior distribution over knowledge, the prior distribution can be utilized to select appropriate knowledge so as to generate informative responses even the actual response is unknown.
Ren \textit{et al.} \cite{ren2019thinking} propose a Global-to-Local Knowledge Selection mechanism using the global perspective to select appropriate background knowledge. A topic transition vector is learned from the dialogue context and external knowledge by a distantly supervised learning schema to select the most likely text fragments. The vector is then used to guide the Local Knowledge Selection module at decoding stage to generate fluency and appropriate responses. Zhao \textit{et al.} \cite{Zhao2020Low-Resource} represent a disentangled response decoder to separate parameters relying on knowledge-grounded dialogues from the whole model to solve the problem of lacking knowledge-grounded training data. The decoder is composed of three components, including \textit{Language Model} to generate common words, \textit{Context Processor} to generate context words, and \textit{Knowledge Processor} to generate words from knowledge document by a hierarchical attention mechanism.

\subsection{Memory network-based knowledge understanding}
After extracting relevant knowledge facts, the most import is to read and understand the textual knowledge to enhance the input representation and guide text generation. Memory network \cite{sukhbaatar2015end} is proposed to improve the poor memory ability of RNN, using external memory component to realize the storage of long-term memory. Because of its powerful memory storage capacity, memory network is widely used in knowledge-enhanced text generation systems to retrieve, read and condition on external knowledge.

Madott \textit{et al.} \cite{madotto2018mem2seq} augment the memory network with a sequential generative architecture. Knowledge facts are fed into memory network to update the input query vector. A GRU is used as a dynamic query generator to generate the output words which will produce two distribution to decide whether to generate common words or memory contents. In order to carefully read and understand the retrieved knowledge, Dinan \textit{et al.} \cite{dinan2018wizard} combine the memory network and Transformer to encode the selected knowledge and the dialogue context to get the higher level semantic representation. Then the dot-product attention between the knowledge and context is performed to retrieved most relevant knowledge for generating the next response.

\subsection{Transformer-based knowledge understanding}
Transformer is an emerging sequential model, which has caused great repercussions in NLP. It exceeds RNN in semantic information abstraction, long-term feature extraction, and task comprehensive feature representation. Many researches have begun to use Transformer to read and attend on external knowledge in text generation systems.

For example, Zhao \textit{et al.} \cite{zhao2019document} make use of multi-head attention mechanism in Transformer to encode the dialogue context, response candidate and the relevant document. Through the hierarchical interaction in the context and document, the importance of different parts of the document and context is determined to select the most appropriate response. Li \textit{et al.} \cite{li2019incremental} employ the multi-head attention to get the vector representation of external knowledge and input. The Incremental Transformer incorporates the vector representation of knowledge and context into the encoding process to encode knowledge utterances span in the multi-turn dialogue. The decoder contains two processes where the first-pass focuses on contextual coherence and the second-pass refines the results of the first-pass by attending on the knowledge to increase the knowledge relevance and correctness. Kim \textit{et al.} \cite{Kim2020Sequential} propose a sequential latent model which sequentially conditions on previously selected knowledge to produce informative responses. The input utterance and knowledge sentences are encoded into vectors, and then model the knowledge selection as latent variables to joint inference knowledge selection of multi-turn dialogue.

\subsection{RL-based knowledge understanding}
Reinforcement learning is an subfield of machine learning that emphasizes how to act based on the state to maximize the expected rewards. Through continuously interacting with the environment, RL can use the rewards and punishments given by the environment to continuously improve the strategy, that is, what kind of actions to take in what kind of state for maximum cumulative rewards. RL is actually very close to the way of human thinking, which is why it is likely to become the future general artificial intelligence paradigm. Based on Deep Q-network (DQN), a typical RL algorithm, Xu \textit{et al.} \cite{xu2019end} propose the knowledge-routed DQN to manage topic transitions during the dialogue. The relational refinement branch encodes relations among different symptoms and the knowledge-routed graph branch decides policy in RL under different medical knowledge. The two branches ensure that the dialogue manager, the agent interacting with the environment, can make more reasonable decisions from knowledge guiding and relation encoding.

\section{Conclusion and Future Directions}
This survey makes a systematic literature review of the research trends of knowledge-enhanced text generation. With the help of external knowledge, text generation system can understand input text more deeply and comprehensively, and generate more informative text, which is a very promising research direction. As an emerging research direction, there are many open issues in the research of knowledge-enhanced text generation system, which will be briefly discussed here.

\subsection{Combining structured and unstructured knowledge}
At present, researches mainly focus on incorporating one form of external knowledge. If the two forms of knowledge are combined, more appropriately and informatively text may be generated. Structured KG can narrow down knowledge candidates using the prior information such as entities and graph paths. Unstructured KB can provide abundant information to enhance text generation but we need strong capability of natural language understanding to select useful information. Both forms of knowledge have their own advantages and disadvantages. Due to their structural differences, it is a challenging research direction to combine the structured and unstructured knowledge into generation systems, which will certainly bring promising progress to the knowledge-enhanced text generation system.

\subsection{Lifelong learning}
We humans continuously learn new knowledge, update our knowledge base to adapt to the fast-changing pace of society. However, existing text generation systems mostly utilize fixed knowledge bases whose knowledge do not keep updating in real time. To make text generation models more anthropomorphic, they should have the ability of continuous lifelong learning. A meaningful exploration of this is discussed by Mazumder et al. \cite{mazumder2018towards}. They propose the lifelong interactive learning and inference model which will actively ask users questions when encountering unknown concepts, and update its knowledge base after corresponding answers are reached. How to continuously obtain information from numerous external inputs and achieve lifelong learning is a important research direction in text generation.
\section*{Acknowledgments}
This work was supported by the National Key R$\&$D Program of China (2017YFB1001800) and the National Natural Science Foundation of China (No. 61772428, 61725205).

\bibliographystyle{named}
\bibliography{Towards_information_rich_logical_text_generation_with_knowledge_enhanced_neural_models}
\end{document}